\documentclass[a4paper]{article}

\usepackage{INTERSPEECH_v2}
\usepackage{multirow}
\usepackage{makecell}

\title{Empirical Evaluation of Parallel Training Algorithms on Acoustic Modeling
}
\name{Wenpeng Li$^1$, Binbin Zhang$^1$, Lei Xie$^{1*}$\thanks{* Corresponding author}, Dong Yu$^2$}
\address{
  $^1$School of Computer Science, Northwestern Polytechnical University, Xi'an, China\\
  $^2$Tencent AI Lab, Seattle, USA}
\email{\{wpli, bbzhang, lxie\}@nwpu-aslp.org, dongyu@ieee.org}

\begin{document}

\newcommand{\tabincell}[2]{\begin{tabular}{@{}#1@{}}#2\end{tabular}}

\maketitle
\begin{abstract}
Deep learning models (DLMs) are state-of-the-art techniques in speech recognition. However, training good DLMs can be time consuming especially for production-size models and corpora. Although several parallel training algorithms have been proposed to improve training efficiency, there is no clear guidance on which one to choose for the task in hand due to lack of systematic and fair comparison among them. In this paper we aim at filling this gap by comparing four popular parallel training algorithms in speech recognition, namely asynchronous stochastic gradient descent (ASGD), blockwise model-update filtering (BMUF), bulk synchronous parallel (BSP) and elastic averaging stochastic gradient descent (EASGD), on 1000-hour LibriSpeech corpora using feed-forward deep neural networks (DNNs) and convolutional, long short-term memory, DNNs (CLDNNs). Based on our experiments, we recommend using BMUF as the top choice to train acoustic models since it is most stable, scales well with number of GPUs, can achieve reproducible results, and in many cases even outperforms single-GPU SGD. ASGD can be used as a substitute in some cases.

\end{abstract}

\vspace{-5pt}
\section{Introduction} \label{sec:intro}

Since 2010, the year in which deep neural networks (DNNs) were successfully applied to the large vocabulary continuous speech recognition (LCVSR) tasks \cite{Dahl2012Context,seide2011conversational, Hinton2012Deep} and led to significant recognition accuracy improvement over the then state of the art, various deep learning models, such as convolutional neural networks (CNNs) \cite{abdel2012applying, abdel2013exploring, Sainath2013Improvements, sainath2013deep, Abdel2014Convolutional, mohamed2014deep}, long short-term memory (LSTM) recurrent neural networks (RNNs) \cite{sak2014long, Ha2014long, Miao2016Simplifying, Ha2015Fast, Sak2015Learning, Senior2015Context, Kai2016Training, zhang2016highway} and their variants \cite{Yu2016Deep, Sun2017, Sercu2016Advances, Zhang2015Feedforward, Zhang2016Compact, Sainath2015Convolutional}, have been developed to further improve the performance of automatic speech recognition (ASR) systems. Albeit achieving the state-of-the-art performance, these deep learning models are time consuming to train well, especially when trained on single-GPU. Trade-offs often need to be made between the scale of model size and training corpora (and thus recognition accuracy) and the training time because even with today's massively parallel GPU it usually takes days or weeks to train large models to desired accuracy on a single GPU.

Many parallel training algorithms have been proposed to speed up training. These algorithms can be  categorized into two classes: model parallelism (e.g., \cite{Dean2012Large, Coates2013Deep}), which exploits and splits the structure of neural networks to distribute computation across GPUs, and data parallelism (e.g., \cite{Dean2012Large, Seide20141, Agarwal2011Distributed}), which splits and distributes data across GPUs to achieve speedup. Model parallelism focuses on computing more parameters at the same time. It allows and is more suitable for training models that are too big to fit in the memory on a single device. On the other hand, data parallelism concentrates on processing more training samples at the same time and is thus best used when there are enormous training samples. In speech recognition, data parallelism is more important since ASR models usually fit well on a single GPU while the training set is often large.

The core problem data parallelism algorithms try to solve is the difficulty in achieving parallelization of mini-batch based stochastic gradient descent (SGD) algorithm \cite{Rumelhart1986Learning}, which is the most popular technique to train deep learning models (DLMs). Several successful techniques, such as asynchronous stochastic gradient descent (ASGD) \cite{Heigold2014Asynchronous, Zhang2013Asynchronous, Paine2013GPU}, blockwise model-update filtering (BMUF) \cite{Chen2016Scalable}, bulk synchronous parallel (BSP) \cite{Valiant1990A, Ma2016Theano, Povey2014Parallel, Su2016Experiments}, 1-bit SGD \cite{Seide20141} and elastic averaging stochastic gradient descent (EASGD) \cite{Zhang2014Deep}, have been proposed recently. Unfortunately, these techniques solve the problem with different assumptions and strategies, have been evaluated only on vastly different data sets and tasks, and there is no theoretical guarantee on their behavior when used to train DLMs. This causes the difficulty in selecting the right parallel algorithm for training models on industrial-size corpora.

In this paper, we evaluate and systematically compare four parallel training algorithms, namely BSP, ASGD, BMUF and EASGD with regard to training speed, convergence behavior, final model's performance, reproducibility, and robustness across models, number of GPUs, and learning control parameters. For all we know, this is the first time these algorithms are compared relatively thoroughly on ASR tasks. It is also the first time EASGD is evaluated for acoustic model training. All the four algorithms were implemented in Kaldi toolkit \cite{Povey2012The} using message passing interface (MPI) for parameter exchange across GPUs. Using the same communication protocol guarantees that the comparison is fair and reliable. To evaluate these algorithms, we train DNNs and CLDNNs \cite{Sainath2015Convolutional} (an architecture that stacks CNNs, LSTMs and DNNs) on 1000hr LibriSpeech \cite{Panayotov2015Librispeech} corpus.

The rest of the paper is organized as follows. In Section \ref{sec:algorithms}, we introduce BSP, ASGD, BMUF and EASGD, and discuss relationships between them. In Section \ref{sec:exp} we describe series of experimental setups and report related results. We conclude the paper in Section \ref{sec:conclusion}.

\vspace{-5pt}
\section{Parallel training algorithms} \label{sec:algorithms}

\subsection{BSP}
The bulk synchronous parallel (BSP) \cite{Valiant1990A} algorithm is often referred to as model averaging. In this model, data are distributed across multiple workers. Each worker updates its local model replica independently using its own portion of data with SGD. Periodically the local models are averaged and the generated global model is synchronized across workers. We denote $w_{t}^{i}$ as the $i$-th worker's local model at time $t$. The global model $\tilde{w}_t$ is computed as 
\begin{equation}
  \tilde{w}_t = \bar{w}_t = \frac{1}{N} \sum_{i=1}^Nw_t^i,
  \label{eq1}
\end{equation}
where $N$ is the number of local workers and $\bar{w}_t$ is the average model of the local models. This algorithm is easy to implement and can achieve linear speedup when communication cost can be ignored (e.g., with large synchronous time) at the cost of recognition accuracy degradation, esp. when the number of workers becomes large.\vspace{-5pt}

\subsection{ASGD}

\begin{figure}[t]
  \centering
  \includegraphics[width=\linewidth]{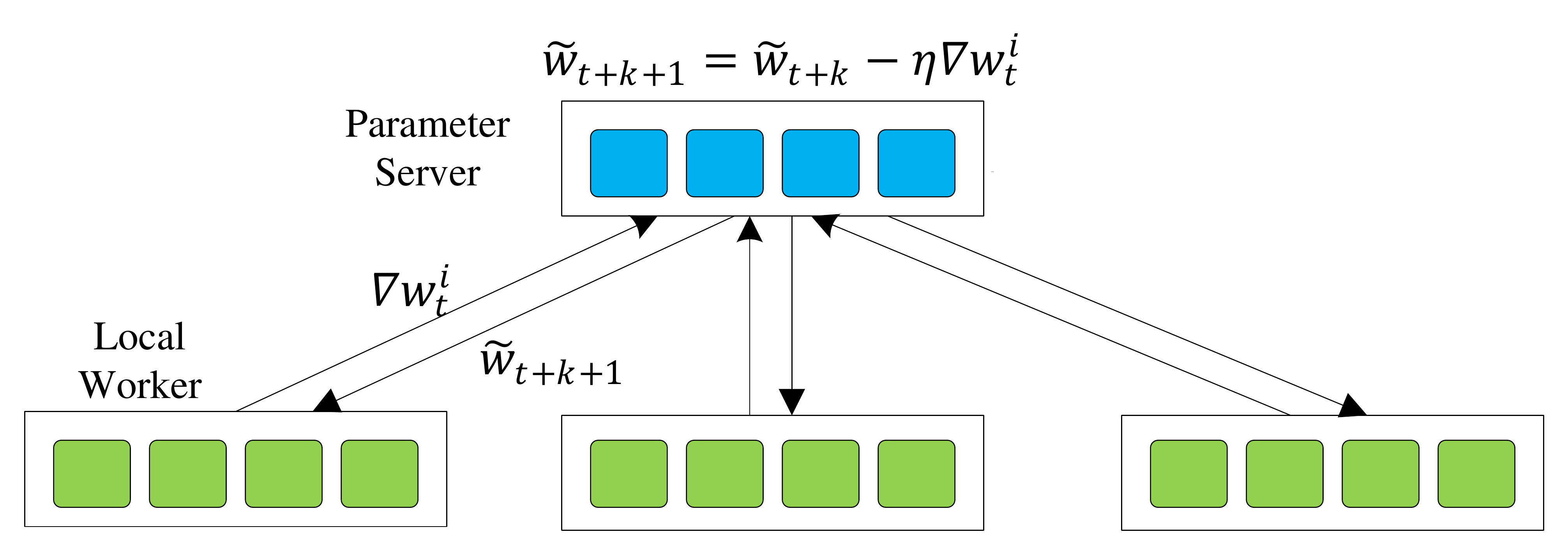}\vspace{-5pt}
  \caption{ASGD architecture. Arrows indicate communication between the parameter server and the workers.}\vspace{-15pt}
  \label{fig:asgd}
\end{figure}

The asynchronous stochastic gradient descent (ASGD) algorithm is the distributed version of SGD. It is proved \cite{Duchi2011Adaptive} that ASGD converges for convex problems. As shown in Figure~\ref{fig:asgd}, ASGD uses a parameter sever and several local workers. Each worker independently and asynchronously pulls the latest global model $\tilde{w}_t$ from the parameter server, computes the gradient $\nabla{w}_t^i$ with a new minibatch, and sends it to the parameter server. The parameter server always keeps the current model. When it receives the gradient $\nabla{w}_t^i$ from worker $i$ it generates the new model 
\vspace{-5pt}
\begin{equation}
  \tilde{w}_{t+k+1} = \tilde{w}_{t+k}-\eta\nabla{w}_t^i \vspace{-3pt}
  \label{eq2}
\end{equation}
where $\eta$ is the learning rate. 

Before worker $i$ sends gradient $\nabla{w}_t^i$ back to parameter server, some other workers may have already added their local gradients to the model and updated the model $k$ times to become $\tilde{w}_{t+k}$. Therefore ASGD essentially adds a ``delayed" gradient $\nabla{w}_t^i$ computed based on the model $\tilde{w}_{t}$ to the model $\tilde{w}_{t+k}$ \cite{Zheng2016Asynchronous}. This may be the reason that ASGD can be unstable: sometimes the model can converge to the same accuracy as that trained with SGD but with more iterations, and sometimes it can never achieve the same performance as SGD, esp. when there are many workers.\vspace{-5pt}

\subsection{BMUF}\vspace{-1pt}

The blockwise model-update filtering (BMUF) algorithm \cite{Chen2016Scalable} can be considered as an improved model averaging technique in which the global model update is implemented as a filter.

In BMUF, the full training set $D$ is partitioned into $M$ non-overlapping blocks and each block is further partitioned into $N$ non-overlapping splits, where $N$ is the number of workers. Each worker updates its local model with its portion of data. The $N$ optimized local models are then averaged using Eq. (\ref{eq1}). Unlike BSP, which treats the average model $\bar{w}_t$ as the global model, BMUF generates the global model $\tilde w_t$ as
\vspace{-3pt}
\begin{equation}
  \tilde w_t = \tilde w_{t-1} + {\Delta}_t, \vspace{-3pt}
  \label{eq3}
\end{equation}
where
\begin{equation}
  {\Delta}_t = \zeta{\Delta}_{t-1} + \eta{G_t}, 0 \leq \zeta < 1, \eta > 0,
  \label{eq4}
\end{equation}
is the global-model update,  
\begin{equation}
  G_t = \bar{w}_t - \tilde w_{t-1}
  \label{eq5}
\end{equation}
is the model-update resulted from a block, $\zeta$ is called block momentum (BM) and $\eta$ is called block learning rate (BLR). We use the formula
\begin{equation}
   \frac {\eta} {N(1-\zeta)} = C
   \label{eq6}
\end{equation}
to set the values of $\zeta$ and $\eta$ empirically, where $C$ is a constant slightly large than 1 and $N$ is the number of workers. Usually, the value of $\eta$ and $C$ both are set to $1.0$ and the value of $\zeta$ is calculated based on Eq. (\ref{eq6}). We implemented CBM-BMUF \cite{Chen2016Scalable} in this work.\vspace{-5pt}

\subsection{EASGD}\vspace{-1pt}

In elastic averaging stochastic gradient descent (EASGD) \cite{Zhang2014Deep}, the loss function is defined as\vspace{-3pt}
\begin{equation}
\label{equation:easgd}
\mathop {\min }\limits_{w_t^1, ..., w_t^N, \tilde w_t} \sum\limits_{i = 1}^N {f({\rm{D|}}{w_t^i})} {\rm{ + }}\frac{\lambda }{2}{\rm{||}}{w_t^i} - \tilde w_t|{|^2}\vspace{-3pt}
\end{equation}
where $D$ is the training set, $f(.)$ is the loss function for local sequential training, $\lambda$ is a hyper-parameter for the quadratic penalty term, $w_t^i$ represents model for the $i$-th worker, and $\tilde w_t$ represents the global model.

From Eq. (\ref{equation:easgd}) we observe that EASGD minimizes the loss summed over all workers, as well as the quadratic difference between the global model and local models. $\frac{\lambda }{2}{\rm{||}}{w_t^i} - \tilde w_t|{|^2}$ is a quadratic regularization term, which forces local workers to stay close to the global model.

By taking the derivative of $w_t^i$ and $\tilde w_t$ in Eq. (\ref{equation:easgd}), we get the update rules for $w_t^i$ and $\tilde w_t$ in synchronous EASGD as\vspace{-3pt}
\begin{equation}
\label{equation:syncdiff}
\begin{array}{l}
{w_{t+1}^i} = {w_t^i} - \eta \nabla {w_t^i} - \eta \lambda ({w_t^i} - \tilde w_t)\\
\tilde w_{t+1} = \tilde w_t - \eta \lambda \sum\limits_{i = 1}^N ( \tilde w_t - {w_t^i})\vspace{-3pt}
\end{array}
\end{equation}
where $\nabla {w_t^i}$ is the stochastic gradient of $f(.)$ with respect to $w_t^i$. 

In asynchronous EASGD, $\nabla {w_t^i}$ is only used in local updating, and the update rules for local  and global models become\vspace{-3pt}
\begin{equation}
\begin{array}{l}
{w_{t+1}^i} = {w_t^i} - \alpha ({w_t^i} - \tilde w_t)\\
\tilde w_{t+1} = \tilde w_t - \alpha (\tilde w_t - {w_t^i})\vspace{-4pt}
\end{array}
\end{equation}
where $\alpha  = \eta \lambda$, which controls the update step for the variable. Small $\alpha$ allows for more exploration as it allows $w^i$ to fluctuate further from $\tilde w$ while large $\alpha$ makes local model perform more exploitation. We only implemented asynchronous EASGD in this work.\vspace{-5pt}

\subsection{Relationships between algorithms}\vspace{-1pt}

These four algorithms, although are different, have relations.

First, ASGD and EASGD are asynchronous algorithms based on the client/server framework, in which the global model is stored on and updated by a parameter server, and each worker computes gradients and updates its local model independently. Workers only exchange parameters with the server and do not communicate with each other. BSP and BMUF, on the other hand, are synchronous algorithms that do not use a server. All workers exchange parameters synchronously with each other.

Second, in ASGD the global model is updated based on the local gradients computed by and sent from workers. In BSP, EASGD and BMUF, however, the global model is a weighted sum of local models instead of gradients.

Third, EASGD and BMUF both introduce extra hyper-parameters whose values may affect the training behavior, while ASGD and BSP have no extra hyper-parameter and thus require less tuning in practice.

Forth, we argue that BMUF actually minimizes the difference \vspace{-3pt}
\begin{equation}
\label{eqvar}
\mathop {\min }\limits_{\tilde w_t} F(\tilde w_t) = \frac{1}{2} \sum_{i=1}^N {\rm{||}}{w_t^i} - \tilde w_t|{|^2}
\end{equation}
between the global and local models. By taking the derivative of $\tilde w$, we get\vspace{-3pt}
\begin{equation}
{\nabla {\tilde w_t}} = \sum_{i=1}^N( \tilde w_t - {w_t^i}).\vspace{-3pt}
\end{equation}
Let $\nabla {\tilde w_t} = 0$. By directly solving $\tilde w_t$, we get\vspace{-3pt}
\begin{equation}
\tilde{w_t} = \frac{1}{N} \sum_{i=1}^N w_t^i.\vspace{-4pt}
\end{equation}
This is the same as BSP in Eq. (\ref{eq1}). If we optimize $\tilde w_t$ using SGD, then\vspace{-5pt}
\begin{equation}
{\tilde w_t} = {\tilde w_{t-1}} - \eta \sum_{i=1}^N(\tilde w_{t-1} - {w_t^i})\vspace{-3pt}
\end{equation}
which is the same as Eq. (\ref{equation:syncdiff}) in EASGD.

Further, if we optimize $\tilde w_t$ using momentum SGD, then\vspace{-5pt}
\begin{equation}
\begin{aligned}
  \tilde w_t &= {\tilde w_{t-1}} - \eta \sum_{i=1}^N(\tilde w_{t-1} - {w_t^i}) - \zeta \nabla {\tilde w_{t-1}} \\
             &= {\tilde w_{t-1}} + \eta^{\prime} G_t +  \zeta^{\prime} \Delta_{t-1} \\
             &= {\tilde w_{t-1}} + \Delta_{t}\vspace{-3pt}
\end{aligned}
\end{equation}
where $\eta^{\prime} = N \eta, \zeta^{\prime} = N \zeta$. 
This is exactly the BMUF update rule in Eq. (\ref{eq3}).

Therefore we conclude that the global model updates of BSP, EASGD and BMUF are derived from the same objective function with different optimization strategies.\vspace{-3pt}

\section{Experiments} \label{sec:exp}

\subsection{Experimental setup}\vspace{-2pt}

In this work, all the models are trained on the 1000hr LibriSpeech \cite{Panayotov2015Librispeech} dataset. The 40-dim FBANK features computed on a 25ms window shifted by 10ms are used. The lexicon and language model (LM) are provided by the dataset. Specifically, the results reported here are all achieved with a full 3-gram LM. We used test-clean and test-other sets for evaluation.

To evaluate the parallel training algorithms, we trained two types of DLMs: DNNs and CLDNNs \cite{Sainath2015Convolutional}. The input to DNNs is the 40-dim FBANK feature with first and second order time derivatives and 11 frame context. The input to CLDNNs is the same as that to DNNs but without the 2nd order time derivatives.  The DNN has 6 hidden layers, each containing 1024 neurons. With 5723 HMM tied-states as output classes, it has about 13.5 million parameters. The CLDNN consists of 1 CNN layer (128 feature maps), 2 DNN layers (1024 neurons) and 2 LSTM layers (1024 memory blocks and 512 projections). With the same output classes as that in DNNs, CLDNNs have about 13.8 million parameters. Both models use the ReLU activation function.

In order to ensure the fairness of the comparison and the credibility of the experimental results, we take the following measures: First, all experiments in this work were carried out on the same computing node with 8 GTX1080 GPUs. Second, the four parallel training algorithms were implemented in the KALDI toolkit. The parameter exchange among GPUs is based on OpenMPI. Third, in all parallel training we used the same initial model which was obtained by one-epoch minibatch-SGD on a single-GPU. Finally, we used the identical learning rate schedule and the same initial learning rate. The learning rate keeps fixed as long as the cross entropy loss on a cross-validation (CV) set decreases by at least 1\%. Then, the learning rate is halved each epoch until the optimization terminates when the cross entropy loss on the CV set decreases by less than 0.1\%.\vspace{-5pt}

\subsection{Experimental results}\vspace{10pt}

\begin{table}[t]
  \caption{WER and training speedup of DNNs trained by single-GPU minibatch-SGD, ASGD, BMUF, BSP, and EASGD. The synchronization period is 5 minibatches for each algorithm.}
  \vspace{-5pt}
  \label{tab:DNN results}
  \centering
   \footnotesize
  \begin{tabular}{|c|c|c|c|c|}
    \hline
    \multirow{2}{*}{\tabincell{c}{Parallel \\ Algorithm}} & \multirow{2}{*}{\tabincell{c}{GPU \\ Number}}
    & \multirow{2}{*}{\tabincell{c}{Training \\ Speedup}} & \multicolumn{2}{c|}{WER(\%)}                     \\
    \cline{4-5}
    \multicolumn{ 1}{|c|}{} & \multicolumn{ 1}{c|}{} & \multicolumn{ 1}{c|}{} & test-clean & test-other      \\
    \hline
    \multirow{2}{*}{ASGD}   & 4                      & 2.74X                  & 5.91       & 15.55           \\
    \cline{2-5}             & 8                      & 4.72X                  & 6.09       & 16.20            \\
    \hline
    \multirow{2}{*}{BMUF}   & 4                      & 2.68X                  & \textbf{5.70}       & \textbf{15.01}      \\
    \cline{2-5}             & 8                      & 4.56X                  & \textbf{5.99}       & \textbf{15.66}       \\
    \hline
    \multirow{2}{*}{BSP}    & 4                      & 2.68X                  & 6.01       & 16.03           \\
    \cline{2-5}             & 8                      & 4.56X                  & 6.21       & 16.55           \\
    \hline
    \multirow{2}{*}{EASGD}  & 4                      & 2.80X                  & 6.04       & 16.02           \\
    \cline{2-5}             & 8                      & 5.00X                  & 6.22       & 16.64           \\
    \hline
    \multirow{2}{*}{\tabincell{c}{Minibatch \\ -SGD}} & \multirow{2}{*}{1}    & \multirow{2}{*}{1.0X}
                                                              & \multirow{2}{*}{5.83} & \multirow{2}{*}{15.44} \\
    \multicolumn{ 1}{|c|}{} & \multicolumn{ 1}{c|}{} & \multicolumn{ 1}{c|}{} & \multicolumn{ 1}{c|}{} & \multicolumn{ 1}{c|}{}      \\
    \hline

  \end{tabular}
\end{table}

\begin{figure}[t]
  \centering
  \includegraphics[width=0.94\columnwidth]{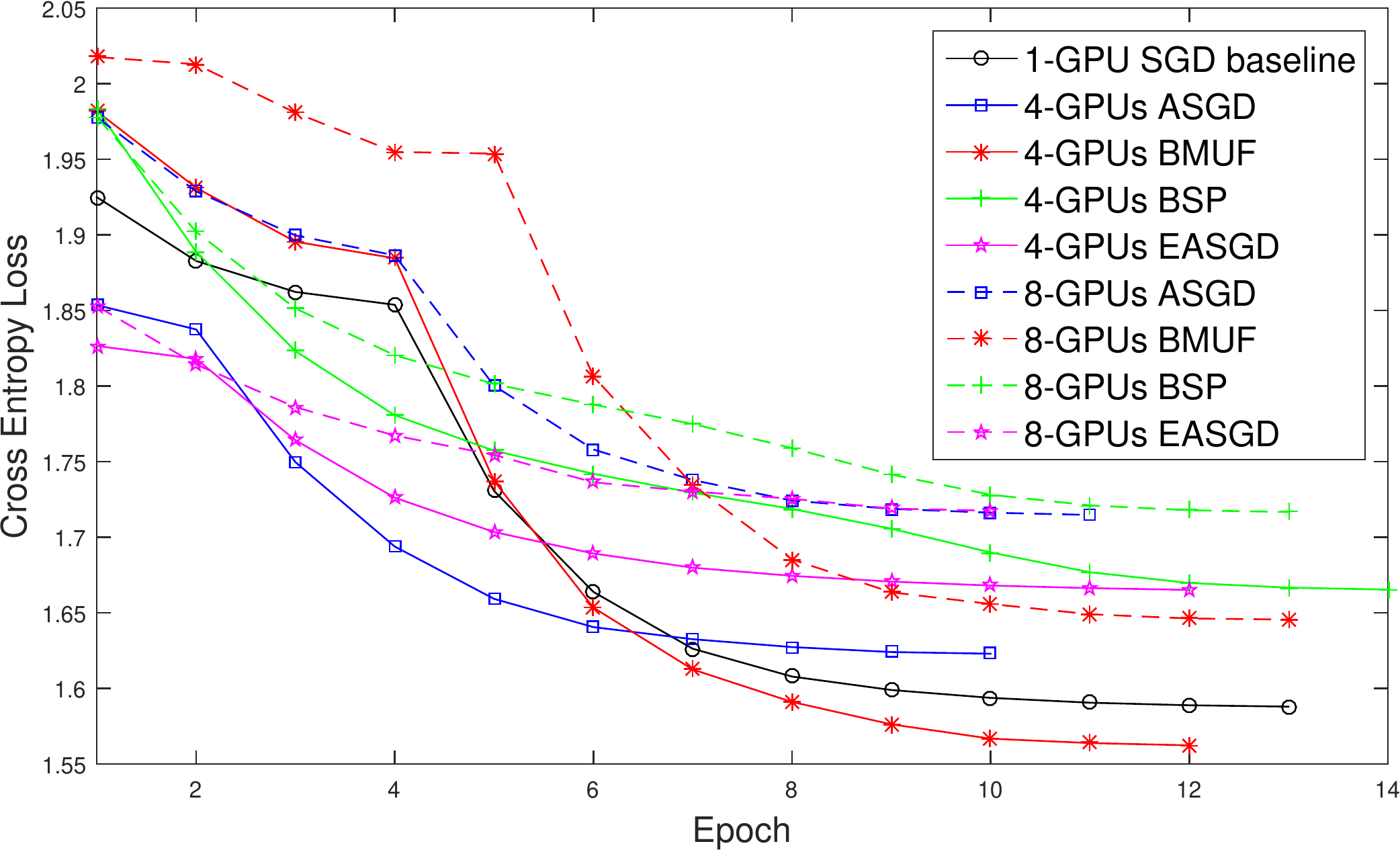}\vspace{-5pt}
  \caption{Learning curves of CE loss on CV set with different algorithms and GPU numbers for DNN training.}
  \label{fig:dnn CV}
\end{figure}

\begin{table}[t]
  \caption{WER and training speedup of DNNs trained by ASGD, BMUF, BSP and EASGD. The most appropriate synchronization period is chosen for each algorithm, respectively. }\vspace{-3pt}
  \label{tab:best sync period}
  \centering
   \footnotesize
  \begin{tabular}{|c|c|c|c|c|}
    \hline
    \multirow{2}{*}{\tabincell{c}{Parallel \\ Algorithm}} & \multirow{2}{*}{\tabincell{c}{Sync \\ Period}}
    & \multirow{2}{*}{\tabincell{c}{Training \\ Speedup}} & \multicolumn{2}{c|}{WER(\%)}                              \\
    \cline{4-5}
    \multicolumn{ 1}{|c|}{} & \multicolumn{ 1}{c|}{} & \multicolumn{ 1}{c|}{} & test-clean          & test-other      \\
    \hline
     ASGD                   & 1                      & 2.22X                  & 5.85                & 15.54           \\
    \hline
     BMUF                   & 80                     & 2.93X                  & \textbf{5.74}       & \textbf{14.87}  \\
    \hline
     BSP                    & 5                      & 2.68X                  & 6.01                & 16.03           \\
    \hline
     EASGD                  & 64                     & 2.99X                  & 6.06                & 15.97           \\
    \hline

  \end{tabular}
\end{table}

\begin{table}[t]
  \caption{WER and training speedup of CLDNNs trained by ASGD, BMUF, BSP and EASGD. The most appropriate synchronization period is chosen for each algorithm, respectively. }\vspace{-3pt}
  \label{tab:CLDNN results}
  \centering
   \footnotesize
  \begin{tabular}{|c|c|c|c|c|}
    \hline
    \multirow{2}{*}{\tabincell{c}{Parallel \\ Algorithm}} & \multirow{2}{*}{\tabincell{c}{GPU \\ Number}}
    & \multirow{2}{*}{\tabincell{c}{Training \\ Speedup}} & \multicolumn{2}{c|}{WER(\%)}                     \\
    \cline{4-5}
    \multicolumn{ 1}{|c|}{} & \multicolumn{ 1}{c|}{} & \multicolumn{ 1}{c|}{} & test-clean & test-other      \\
    \hline
    \multirow{2}{*}{ASGD}   & 4                      & 3.42X                  & 5.37       & 14.08           \\
    \cline{2-5}             & 8                      & 6.11X                  & \textbf{5.48}       & \textbf{14.29}            \\
    \hline
    \multirow{2}{*}{BMUF}   & 4                      & 3.84X                  & \textbf{5.26}       & \textbf{13.80}      \\
    \cline{2-5}             & 8                      & 6.88X                  & 5.63       & 14.65       \\
    \hline
    \multirow{2}{*}{BSP}    & 4                      & 3.50X                  & 5.43       & 14.08           \\
    \cline{2-5}             & 8                      & 6.03X                  & 5.64       & 14.74           \\
    \hline
    \multirow{2}{*}{EASGD}  & 4                      & 3.53X                  & 5.44       & 14.31           \\
    \cline{2-5}             & 8                      & 7.45X                  & 5.55       & 14.54           \\
    \hline
    \multirow{2}{*}{\tabincell{c}{Minibatch \\ -SGD}} & \multirow{2}{*}{1}    & \multirow{2}{*}{1.0X}
                                                              & \multirow{2}{*}{5.26} & \multirow{2}{*}{13.76} \\
    \multicolumn{ 1}{|c|}{} & \multicolumn{ 1}{c|}{} & \multicolumn{ 1}{c|}{} & \multicolumn{ 1}{c|}{} & \multicolumn{ 1}{c|}{}      \\
    \hline

  \end{tabular}
\end{table}
\vspace{-10pt}
\begin{table}[t]
  \caption{WER and training speedup of DNNs trained by ASGD and BMUF on 4 GPUs with various synchronization periods. }\vspace{-3pt}
  \label{tab:sync period results}
  \centering
   \footnotesize
  \begin{tabular}{|c|c|c|c|c|}
    \hline
    \multirow{2}{*}{\tabincell{c}{Parallel \\ Algorithm}} & \multirow{2}{*}{\tabincell{c}{Sync \\ Period}}
    & \multirow{2}{*}{\tabincell{c}{Training \\ Speedup}} & \multicolumn{2}{c|}{WER(\%)}                     \\
    \cline{4-5}
    \multicolumn{ 1}{|c|}{} & \multicolumn{ 1}{c|}{} & \multicolumn{ 1}{c|}{} & test-clean & test-other      \\
    \hline
    \multirow{3}{*}{ASGD}   & 5                      & 2.74X                  & 5.91       & 15.55           \\
    \cline{2-5}             & 20                     & 2.96X                  & 5.97       & 15.78           \\
    \cline{2-5}             & 80                     & \multicolumn{3}{c|}{divergence}                        \\
    \hline
    \multirow{3}{*}{BMUF}   & 5                      & 2.68X                  & 5.70       & 15.01           \\
    \cline{2-5}             & 20                     & 2.90X                  & 5.73       & 14.86           \\
    \cline{2-5}             & 80                     & 2.93X                  & 5.74       & 14.87           \\
    \hline
    \multirow{2}{*}{\tabincell{c}{Minibatch \\ -SGD }} & \multirow{2}{*}{1}    & \multirow{2}{*}{1.0X}
                                                              & \multirow{2}{*}{5.83} & \multirow{2}{*}{15.44} \\
    \multicolumn{ 1}{|c|}{} & \multicolumn{ 1}{c|}{} & \multicolumn{ 1}{c|}{} & \multicolumn{ 1}{c|}{} & \multicolumn{ 1}{c|}{}      \\
    \hline

  \end{tabular}
\end{table}

\begin{table}[t]
  \caption{WER and training speedup of DNNs trained by BMUF on 4 GPUs with various minibatch sizes.} \vspace{-3pt}
  \label{tab:minibatch results}
  \centering
  \footnotesize
  \begin{tabular}{|c|c|c|c|c|}
    \hline
    \multirow{2}{*}{\tabincell{c}{Minibatch \\ Size}} & \multirow{2}{*}{\tabincell{c}{Sync \\ Period}}
    & \multirow{2}{*}{\tabincell{c}{Training \\ Speedup}} & \multicolumn{2}{c|}{WER(\%)}                     \\
    \cline{4-5}
    \multicolumn{ 1}{|c|}{} & \multicolumn{ 1}{c|}{} & \multicolumn{ 1}{c|}{} & test-clean & test-other      \\
    \hline
    \multirow{4}{*}{256}    & Single-GPU             & 1.0X           & 5.80       & 15.13           \\
    \cline{2-5}             & 5                      & 1.85X          & 5.86       & 15.31           \\
    \cline{2-5}             & 20                     & 2.54X          & 5.86       & 15.26           \\
    \cline{2-5}             & 80                     & 3.21X          & 5.87       & 15.13           \\
    \hline
    \multirow{4}{*}{1024}   & Single-GPU             & 1.0X           & 5.72       & 14.80           \\
    \cline{2-5}             & 5                      & 2.14X          & 5.71       & 15.02           \\
    \cline{2-5}             & 20                     & 2.76X          & 5.74       & 15.02           \\
    \cline{2-5}             & 80                     & 2.98X          & 5.69       & 14.90           \\
    \hline
    \multirow{4}{*}{4096}   & Single-GPU             & 1.0X           & 5.83       & 15.44           \\
    \cline{2-5}             & 5                      & 2.68X          & 5.70       & 15.01           \\
    \cline{2-5}             & 20                     & 2.90X          & 5.73       & 14.86           \\
    \cline{2-5}             & 80                     & 2.93X          & 5.74       & 14.87           \\
    \hline

  \end{tabular}
\end{table}

\subsubsection{DNN results}
In DNN training, the minibatch size was set to 4096. Table~\ref{tab:DNN results} compares training speedups and word error rate (WER) on testing sets with four parallel training algorithms on 4 GPUs and 8 GPUs. When using same synchronization period, EASGD achieved the best training speedup, followed by ASGD, BMUF and BSP. However, BMUF achieved the best WER in both 4 and 8 GPU cases, and even outperformed the single-GPU SGD. As Figure~\ref{fig:dnn CV} shows, the convergence trend of BMUF is similar to minibatch-SGD with single-GPU in the CV set.  To further verify our conclusions, we chose the most appropriate synchronization period for each algorithm based on the literature. The results in Table~\ref{tab:best sync period} show that BMUF still performed the best.\vspace{-6pt}

\subsubsection{CLDNN results}\vspace{-2pt}

In CLDNN training, we computed gradients on 100 subsequences from different utterances in the same time. The truncated BPTT with truncation step of 20 was used to train the models. The appropriate synchronization period was chosen for each algorithm based on the literature. Specifically, the synchronization periods for BSP, ASGD, BMUF, and EASGD are 5, 5, 80, and 64 minbatches, respectively. Table~\ref{tab:CLDNN results} shows that BMUF achieved the best WER and training speedup with 4 GPUs and ASGD achieved the best WER with 8 GPUs.\vspace{-6pt}

\subsubsection{Synchronization period}\vspace{-2pt}

The choice of synchronization period $\tau$ affects the behavior of BMUF and ASGD. As Table~\ref{tab:sync period results} shows, the WER of ASGD gradually increased and the training even diverged when increasing $\tau$. This is because ASGD suffers from the problem of delayed gradient update and larger synchronization period results in greater latency. In contrast, we observed no performance degradation on BMUF when varying the synchronization period. As for training speedup, when synchronization period $\tau$ is small, the parameter exchange among multi-GPUs is quite frequent and the communication overhead is the main bottleneck of training speed. So as $\tau$ increases (from 5 to 20 minibatches in Table~\ref{tab:sync period results}), the communication overhead decreases and training speedup increases. When the value of $\tau$ continues to increase, the communication overhead is reduced so that the computation speed becomes the main bottleneck. Therefore the training speedup almost keeps unchanged as the synchronization period increases (from 20 to 80 minibatches in Table~\ref{tab:sync period results}).\vspace{-5pt}

\subsubsection{Minibatch size}\vspace{-2pt}

Table~\ref{tab:minibatch results} compares three different minibatch sizes in BMUF. With single-GPU training, the training speed is lower when smaller minibatch size is used. This is because with small minibatch size the GPU power is not fully utilized and the model is updated more frequently.

The training speedup $s$ of multi-GPU training is calculated as\vspace{-5pt}
\begin{equation}
   s = \frac {t_s} {f(t_s, N) + t_c}\vspace{-3pt}
   \label{speedup}
\end{equation}
where $t_s$ is the computation time through one epoch of dataset on single-GPU, $N$ is the number of GPUs, $t_c$ is the communication overhead, and \vspace{-6pt}
\begin{equation}
   f(t_s, N) = \alpha \frac {t_s} {N}\vspace{-3pt}
   \label{f_n}
\end{equation}
 is a decreasing function over $N$. When the minibatch can fill the GPU $\alpha$ is 1, otherwise $\alpha$ is greater than 1. According to Eqs. (\ref{speedup}) and (\ref{f_n}), we get\vspace{-5pt}
\begin{equation} 
   s = \frac {1} {\frac {\alpha} {N} + \frac {t_c} {t_s}}. \vspace{-5pt}
   \label{speedup simple}
\end{equation}
This means the training speedup $s$ of multi-GPU parallel training depends on $\frac {t_c}{t_s}$. When synchronization period $\tau$ is small, the communication overhead $t_c$ is large due to frequent parameter exchange among GPUs. Although $t_c$ and $t_s$ decrease with the increase of minibatch size, $t_c$ decreases more quickly. Therefore the larger minibatch size leads to the greater training speedup. When $\tau$ is large, however, $t_c$ is small.  $t_s$ decreases more quickly with the increase of minibatch size and causes decreasing in training speedup. However, from the perspective of absolute training speed, it benefits from the growth of minibatch size. Eq.~(\ref{speedup simple}) also explains why the speedup of CLDNN (Table~\ref{tab:CLDNN results}) is much larger than DNN (Table~\ref{tab:DNN results}). In the same GPU number and synchronization period, $t_c$ of DNN and CLDNN is similar, but $t_s$ of CLDNN is larger than DNN .\vspace{-5pt}

\section{Conclusions} \label{sec:conclusion}\vspace{-2pt}

We implemented four parallel training algorithms, discussed the relationship among them, and evaluated them on speech recognition tasks. The experimental results show that BMUF and ASGD consistently outperform BSP and EASGD. BMUF, in particular, achieved the best performance without frequent synchronization, and even outperformed the single-GPU SGD in some cases. We conjecture that the momentum used in BMUF global model update makes the global model not only related to each local model but also the previous global model. ASGD also achieved pretty good performance and the lager training speedup with the same synchronization period. Profit fully from the asynchronous properties, ASGD is insensitive to the difference of computing capacity of workers, but sensitive to the synchronization period and suffers from the poor reproducibility.\vspace{-5pt}

\section{Acknowledgements} \label{sec:acknowledgements}

This work is supported by the National Natural Science Foundation of China (Grant No. 61571363).

\bibliographystyle{IEEEtran}

\bibliography{parallel-training-interspeech2017}


\end{document}